\documentclass[lettersize,journal]{IEEEtran}
\usepackage{amsmath,amsfonts}
\usepackage{algorithmic}
\usepackage{algorithm}
\usepackage{array}
\usepackage[caption=false,font=footnotesize,labelfont=rm,textfont=rm]{subfig}
\usepackage{textcomp}
\usepackage{stfloats}
\usepackage{url}
\usepackage{verbatim}
\usepackage{graphicx}
\usepackage{cite}
\usepackage{booktabs}
\usepackage{multirow}
\usepackage{colortbl}
\usepackage{xcolor}
\usepackage{makecell}
\usepackage{pifont}
\usepackage{arydshln}
\begin{document}

\title{Vision-Language Meets the Skeleton: Progressively Distillation with Cross-Modal Knowledge for 3D Action Representation Learning}

\author{Yang Chen, 
Tian He, 
Junfeng Fu, 
Ling Wang,~\IEEEmembership{Member,~IEEE},
\\Jingcai Guo,~\IEEEmembership{Member,~IEEE},
Ting Hu, Hong Cheng,~\IEEEmembership{Senior Member,~IEEE}

\thanks{This work was supported by the National Key Research and Development Program of China (No.2022YFE0133100), the Hong Kong RGC General Research Fund (No. 152211/23E and 15216424/24E), the National Natural Science Foundation of China (No. 62102327), and PolyU Internal Fund (No. P0043932). This research was also supported by NVIDIA AI Technology Center (NVAITC).(\textit{Corresponding author: Ling Wang.})}
\thanks{Yang Chen, Junfeng Fu and Ling Wang are with the School of Information and Communication Engineering, University of Electronic Science and Technology of China, Chengdu, China (email: csychen@std.uestc.edu.cn; junfengfu@std.uestc.edu.cn; eewangling@uestc.edu.cn).}
\thanks{Tian He and Hong Cheng are with the School of Automation Engineering, University of Electronic Science and Technology of China, Chengdu, China (email: tianhe@std.uestc.edu.cn; hcheng@uestc.edu.cn).}
\thanks{Jingcai Guo is with the Department of Computing, The Hong Kong Polytechnic University, Hong Kong SAR, China. Meanwhile, he is also with Hong Kong Polytechnic University Shenzhen Research Institute, Shenzhen, China (email: jc-jingcai.guo@polyu.edu.hk).}
\thanks{Ting Hu is with the No.1 Orthopedic Hospital of Chengdu, Chengdu, China (email: chinaht2008@126.com).}
}

\markboth{IEEE~TRANSACTIONS~ON~MULTIMEDIA, 2024}%
{IEEE TRANSACTIONS ON MULTIMEDIA, 2024}


\maketitle

\begin{abstract}
Skeleton-based action representation learning aims to interpret and understand human behaviors by encoding the skeleton sequences, which can be categorized into two primary training paradigms: supervised learning and self-supervised learning.
However, the former one-hot classification requires labor-intensive predefined action categories annotations, while the latter involves skeleton transformations (e.g., cropping) in the pretext tasks that may impair the skeleton structure. To address these challenges, we introduce a novel skeleton-based training framework (C$^2$VL) based on \textit{C}ross-modal \textit{C}ontrastive learning that uses the progressive distillation to learn task-agnostic human skeleton action representation from the \textit{V}ision-\textit{L}anguage knowledge prompts. Specifically, we establish the vision-language action concept space through vision-language knowledge prompts generated by pre-trained large multimodal models (LMMs), which enrich the fine-grained details that the skeleton action space lacks. Moreover, we propose the intra-modal self-similarity and inter-modal cross-consistency softened targets in the cross-modal representation learning process to progressively control and guide the degree of pulling vision-language knowledge prompts and corresponding skeletons closer. These soft instance discrimination and self-knowledge distillation strategies contribute to the learning of better skeleton-based action representations from the noisy skeleton-vision-language pairs. During the inference phase, our method requires only the skeleton data as the input for action recognition and no longer for vision-language prompts. 
Extensive experiments on NTU RGB+D 60, NTU RGB+D 120, and PKU-MMD datasets demonstrate that our method outperforms the previous methods and achieves state-of-the-art results. Code is available at: \url{https://github.com/cseeyangchen/C2VL}.
\end{abstract}

\begin{IEEEkeywords}
Action Recognition, Vision-Language, Cross-Modal, Self-Supervised, Contrastive Learning, LMMs.
\end{IEEEkeywords}

\section{Introduction}
\IEEEPARstart{A}{ction} recognition has become a hot topic due to its wide range of application fields, such as human-computer interaction, sports analysis, and dangerous behavior warning \cite{sato2023prompt}, etc. 
In the past ten years, the continuous development of depth sensor cameras like Kinect \cite{zhang2012microsoft} has facilitated the easy acquisition of the human body's skeleton (3D joints coordinate).
Compared with RGB-D sequences, the 3D skeleton is naturally robust to light, background noise, and viewpoint changes while also having the advantages of being lightweight and privacy-preserving.

\begin{figure}[t]
  \centering
   \includegraphics[width=0.9\linewidth, height=0.9\linewidth]{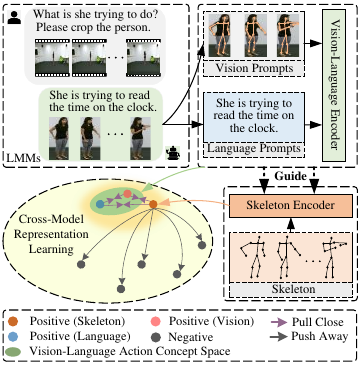}
   \caption{Our proposed approach (C$^2$VL) utilizes the vision-language knowledge prompts as supervision to learn task-agnostic 3D human action representations.}
   \label{fig:simple_framework}
\end{figure}

Skeleton-based action recognition methods are primarily categorized into two distinct paradigms: supervised learning and self-supervised learning. In the former, skeleton data serves as the input, with human-designed action category labels represented as one-hot vectors employed for supervision to design complex network architectures \cite{chen2021channel, chi2022infogcn,Lee_2023_ICCV}. This training process aims to learn task-specific spatial-temporal representations for actions, which has achieved impressive performance.  On the other hand, self-supervised learning approaches generate positive-negative pairs by designing pretext tasks of data transformations (rotation, masking, and cropping, etc.) to explore task-agnostic action representation in a label-free manner based on contrastive learning  \cite{li20213d,zhu2023modeling,lin2023actionlet}. These methods leverage pretext tasks for unsupervised pre-training and then fine-tuning them for downstream tasks. The results suggest that the self-supervised approaches have the potential to surpass supervised learning methods.

However, while the aforementioned methods have proven effective in representing human actions, several challenges remain. (1) The one-hot vector classification in supervised learning requires time-consuming and labor-intensive annotating of numerous labels while restricting the generalization performance of these methods. (2) For self-supervised learning, data transformations (rotation, masking, cropping, etc.) of the pretext task may impair the structural information inherent in skeleton action, leading to information bias. For example, the motion of the "hand rising" action will be impaired if the masking transformation is applied to the hand joints \cite{lin2023actionlet}. Inspired by the recent success in multi-modal contrastive learning \cite{radford2021learning, wang2021actionclip, xiang2023generative}, we consider designing a novel skeleton training paradigm guided by cross-modal representation learning with auto-generated vision-language knowledge prompts, which can boost the task-agnostic generalization capacity while obviating numerous human-effort annotation and irreversible skeleton impairments, as shown in Fig. \ref{fig:simple_framework}.

Specifically, our approach leverages sample-specific vision-language knowledge prompts as the supervision to guide the skeleton encoder in learning task-agnostic action representation, which is called C$^2$VL. 
Firstly, we utilize two LMMs (Grounding DINO \cite{liu2023grounding} and LLaVA  \cite{liu2023llava}) as the vision and language engines, respectively. The former is an open-set object detector, and the latter is a foundation language model. Then, we employ text prompts and visual questions for them to auto-generate one-to-one vision knowledge prompts related to human action images and one-to-one language knowledge prompts corresponding to action descriptions, respectively.
Subsequently, the vision-language encoders extract features from these knowledge prompts to establish the vision-language action concept space, enriching object-related details and fine-grained descriptions absent in the skeleton action space. However, these one-to-one format skeleton-knowledge pairs exhibit many-to-many correspondences among different samples because LMMs generate similar semantic descriptions for different skeleton actions, which causes the noisy skeleton-vision-language joint action space. Therefore, we introduce the intra-modal self-similarity and inter-modal cross-consistency softened targets in the cross-modal self-knowledge distillation learning to progressively control and guide the degree of bringing vision-language knowledge prompts and corresponding skeletons closer in the noisy joint space. This soft contrastive learning process effectively guides the skeleton encoder to learn better task-agnostic 3D human action representation than the hard alignment in CLIP methods. During the inference phase, only skeleton data is needed as the input for action recognition, and vision-language knowledge prompts are omitted without incurring additional computational costs.

Our contributions can be summarized as follows:
\begin{itemize}
    \item To the best of our knowledge, this is the first work that utilizes vision-language knowledge prompts as supervision to learn task-agnostic skeleton-based action representations in the training phase. During the inference phase, only the skeleton data is used as the input for skeleton-based action recognition.
    \item We employ LMMs as the offline vision-language engines to generate vision-language knowledge prompts concerning action descriptions to establish action concept space, which enriches object-related and fine-grained details that the skeleton action space lacked.  
    \item We introduce the intra-modal self-similarity and inter-modal cross-consistency softened targets to relax the strict one-to-one constraint, effectively guiding the instance discrimination and knowledge distillation process in the noisy joint action space.
    \item Extensive experiments show the superior performance of our training paradigm on the NTU RGB+D 60, NTU RGB+D 120, and PKU-MMD datasets. Notably, our self-supervised method even surpasses some supervised learning methods and has a significant margin improvement under semi-supervised learning scenarios.
\end{itemize}

\section{Related Work}
\label{sec:relatedwork}
In this section, we mainly introduce the related work of skeleton-based action recognition, multi-modal representation learning, and vision-language pretraining with noisy data.

\subsection{Skeleton-based Action Recognition Methods}
\label{sec:skeleton-based action recognition}
Supervised skeleton-based action recognition has been the dominant paradigm. CNN-based approaches \cite{wang2016action, liu2017enhanced} transform skeleton data into image-like structures, and RNN-based methods \cite{liu2016spatio, zhang2017view} delve into long-term contextual relationships from joint sequences, both of them ignore the skeleton's inherent graph topology. GCN-based methods \cite{chen2021channel,song2022constructing,chi2022infogcn} regard human joints and bones as graph nodes and edges, respectively, enabling effective exploration of the spatial-temporal representation of actions while keeping structural properties. However, the above-mentioned supervised approaches have the disadvantage of requiring extensive human-designed category annotations. Much research pays attention to self-supervised approaches, which are adept at learning task-agnostic representation in a label-free manner. Reconstruction and contrastive learning are two main problem-solving concepts. They reconstruct skeleton structures by masked modeling \cite{yang2021skeleton, wu2023skeletonmae,zeng2023contrastive,xu2021prototypical} or generate positive-negative pairs \cite{lin2023actionlet,zhu2023modeling,li20213d,wang2023learning,wang2024localized} through skeleton transformation  (e.g., rotation, cropping) respectively to explore action representation. Unfortunately, these pretext tasks impair the topological structure of the skeleton, leading to information bias \cite{lin2023actionlet}. [\textit{Summary}]: In this paper, we first explore the possibility of learning action representation for skeleton-based action recognition from other modalities (vision and language) without
label annotation or skeleton damage.

\subsection{Multimodal Representation Learning Methods}
\label{sec:multi-modal learning}
Numerous multimodal approaches have demonstrated the capacity to learn powerful representations from different modalities \cite{dong2022co}. Inspired by their success, foundational vision-language models such as CLIP \cite{radford2021learning} and ALIGN \cite{jia2021scaling} have gained widespread adoption in the domain of RGB-based action recognition \cite{aganian2023object, sato2023prompt,mondal2023actor,wang2021actionclip,zheng2023actionprompt,tevet2022motionclip,wu2023bidirectional,kalakonda2023action,xiang2023generative}. Many approaches rely on human-designed coarse-grained category names for video action classification \cite{wang2021actionclip}, pose estimation \cite{zheng2023actionprompt} and action generation \cite{tevet2022motionclip}, etc. PalmCohashNet utilizes multimodal co-learning hash technology for identity authentication \cite{dong2022co}.
Much recent research gradually employs LLMs (GPT-3, etc.) to generate detailed action descriptions of category names for RGB-based action classification \cite{wu2023bidirectional} and generation \cite{kalakonda2023action}. GAP \cite{xiang2023generative} is the first work to utilize LLMs (GPT-3) for skeleton-based action recognition. Regrettably, although these approaches enrich category names with meaningful semantics, the relationship between each action description and skeleton sample falls under the one-to-many format, which belongs to the supervised learning paradigm. [\textit{Summary}]: In contrast, our proposed framework differs in two aspects. Firstly, we leverage LMMs (Grounding DINO \cite{liu2023grounding} and LLaVA \cite{liu2023llava}) as offline engines to generate sample-level vision-language knowledge prompts in a one-to-one format without requiring labor-intensive annotations. Second and more importantly, we utilize these one-to-one vision-language knowledge prompts as the supervisory signal to learn skeleton representation in a self-supervised learning paradigm.

\subsection{Contrastive Learning with Noisy Pairs Methods}
\label{sec:noisy knowledge distillation}
Contrastive learning is the core of the recent multimodal self-supervised learning methods, which pulls the positive pairs closer and pushes the negative pairs away \cite{li2021learning,radenovic2023filtering}. However, the noisy many-to-many correspondences existed in the label-intensive annotated and web-collected pairs, showing the potential semantic similarity between negative samples. This phenomenon leads to the computational inefficiency of the conventional contrast process that pursues the hard alignment \cite{andonian2022robust}. Therefore, several techniques have been developed from the soft alignment perspective to learn robust representations from noisy pairs \cite{srinivasa2023cwcl,morgado2021robust,andonian2022robust,gao2024softclip,huang2024cross,guo2022contrastive}, including the weighting mechanism \cite{srinivasa2023cwcl,morgado2021robust}, k-nearest neighbors scheme \cite{guo2022contrastive}, and knowledge distillation as softening targets \cite{andonian2022robust,gao2024softclip,huang2024cross}. [\textit{Summary}]: In this paper, we design two softened targets with the self-distillation mechanism to progressively mitigate the adverse effects of noisy pairs and refine inconsistent alignment for more coherent representations.

\section{Method}
\label{sec:method}

\begin{figure*}[t]
  \centering
   \includegraphics[width=0.95\linewidth]{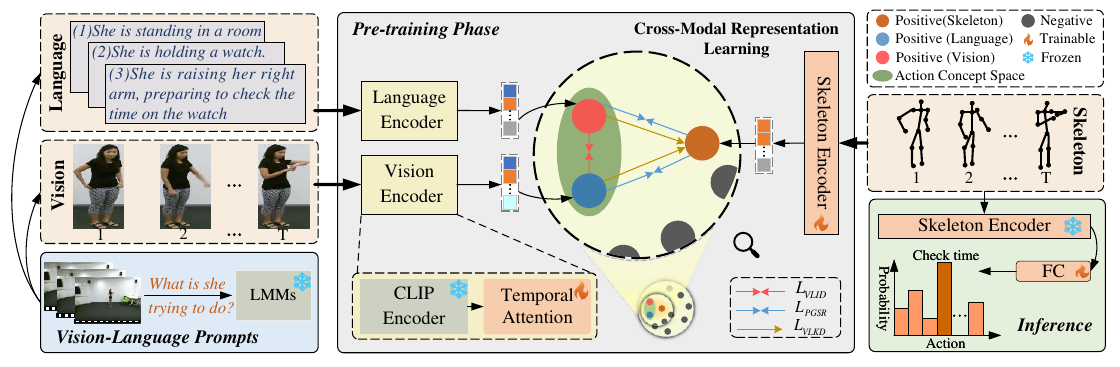}
   \caption{The pipeline of our proposed approach. Before the pre-training phase, the vision and language knowledge prompts are generated regarding skeleton sequences by offline LMMs (Grounding DINO \cite{liu2023grounding} and LLaVA \cite{liu2023llava}) with text prompts and visual questions. In the pre-training phase, the skeleton data is utilized as the input for the skeleton encoder to learn action representation in the skeleton action space. The vision encoder and language encoder are employed to extract features from vision and language knowledge prompts, contributing to the creation of the vision-language action concept space that enhances fine-grained details not captured in the skeleton space. Subsequently, the degree to which pairs consisting of vision-language knowledge prompts and their corresponding skeleton are brought closer should be progressively guided by the intra-modal self-similarity and inter-modal cross-consistency softened targets. During the inference phase, we only utilize the former pre-trained skeleton encoder with a fully connected layer and skeleton data for skeleton-based action recognition without vision-language knowledge prompts.}
   \label{fig: framework}
\end{figure*}

In this section, we begin with an introduction to generating vision-language knowledge prompts (Section \ref{sec: knowledge}). Following that, we describe the contrastive learning process with InfoNCE loss (Section \ref{sec: infonce}). Then, we introduce two soft targets (Section \ref{sec: self-distillation}). Finally, the progressive training objectives with dynamic mechanisms are described (Section \ref{sec: progressive}). The pipeline of the method is shown in Fig. \ref{fig: framework}.

\subsection{Vision-Language Action Concept Space}
\label{sec: knowledge}
\begin{figure}[t]
  \centering
   \includegraphics[width=0.93\linewidth, height=0.4\linewidth]{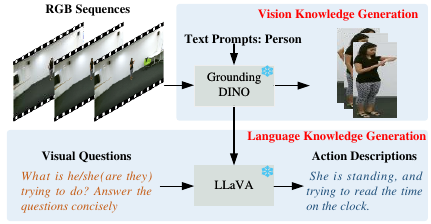}
   \caption{The establishment of vision-language action concept space.}
   \label{fig: knowledge}
\end{figure}

For a given skeleton dataset $D \in\left \{S_{i} \right \}_{i=1}^{N}$, we first generate the sample-level vision-language knowledge prompts $\left \{ (V_{i}, L_{i}) \right \}_{i=1}^{N}$ by offline LMMs (Grounding DINO \cite{liu2023grounding} and LLaVA \cite{liu2023llava}), where $N$ denotes the number of samples in dataset. $S_{i}$ is the $i$-th skeleton sample, $V_{i}$ is the $i$-th vision sample, and $L_{i}$ is the $i$-th language sample.
These sample-level vision-language knowledge prompts have a one-to-one corresponding with skeleton sequence, which establishes the vision-language action concept space as shown in Fig. \ref{fig: knowledge}. 


\subsubsection{Vision Knowledge Prompts}
\label{sec: vision_knowledge}
To the best of our knowledge, prior skeleton-based action recognition methods have yet to explore the possibility of utilizing videos to guide the learning of skeleton representation. Here, we introduce the video modality to enrich the motion patterns and object-related details of skeleton representation. These raw videos and skeleton sequences have one-to-one correspondence as they are captured simultaneously by Kinect devices. However, raw videos are challenged to capture fine-grained details and subtle movements because of interference from irrelevant background environments. To mitigate this issue, we leverage a pre-trained open-set object detector called Grounding DINO \cite{liu2023grounding}, with text prompts ``\textit{\textbf{person}}'' to crop the human action images as vision knowledge prompts precisely. These vision knowledge prompts only consist of human action regions enhancing the motion patterns in the vision action concept space.   

\subsubsection{Language Knowledge Prompts}
\label{sec: language_knowledge}
Inspired by the success of CLIP in handling image-text pairs, numerous methods \cite{wang2021actionclip,zheng2023actionprompt,tevet2022motionclip,xiang2023generative} employ human-designed action names and LLMs-generated movement descriptions for action recognition. However, this language information is at the category level and has a one-to-many relationship with the skeleton sequences, so more sample-specific detail descriptions are needed. Therefore, we aim to establish sample-level language action descriptions to enrich the action concept space without requiring expensive annotation and labor-intensive effort. Fortunately, LMMs have an advantage in visual question answering, presenting a new opportunity. Here, we utilize LLaVA \cite{liu2023llava} with suitable question prompts “\textit{\textbf{Is he/she or are they holding anything in the hand? Is he/she or are they standing or sitting? What is he/she or are they trying to do? Answer the questions concisely}}” to comprehend and generate action descriptions to the vision action images autonomously. These language knowledge prompts provide sample-specific fine-grained descriptions in the language action concept space.

\subsection{Contrastive Learning with InfoNCE Loss}
\label{sec: infonce}
Consider a batch of $B$ paired skeleton-vision-language tuples $\left \{ (S_{i}, V_{i}, L_{i}) \right \}_{i=1}^{B}$ drawn from the above-mentioned multi-modal space, we aim to learn skeleton encoders $E_{s}$ based on cross-modal contrastive learning. Specifically, for the $i$-th pair, the skeleton data $S_{i}$, vision data $V_{i}$ and language data $L_{i}$ are input into the skeleton encoder $E_{s}$, vision encoder $E_{v}$ and language encoder $E_{t}$ respectively to get the corresponding modality features. Then, various projectors map these features to a shared latent space for generating L2-normalized embedding pairs $\left \{ (\mathrm{\mathbf{s}}_{i},\mathrm{\mathbf{v}}_{i},\mathrm{\mathbf{l}}_{i}) \right \}_{i=1}^{\mathrm{B}}$, where $\mathrm{\mathbf{s}}_{i},\mathrm{\mathbf{v}}_{i},\mathrm{\mathbf{l}}_{i}\in \mathbb{R}^{d}$. We use the InfoNCE \cite{oord2018representation} loss to align the skeleton and vision spaces, which focuses on maximizing the similarity between $\mathrm{\mathbf{s}}_{i}$ and $\mathrm{\mathbf{v}}_{i}$ to pulls the paired skeleton-vision embeddings together in two directions. The loss $\mathcal{L}_{s2v}$ for aligning skeleton to vision is defined as follows:
\begin{equation}
    \mathcal{L}_{s2v}=-\frac{1}{\mathrm{B}}\sum\limits_{i=1}^{\mathrm{B}}\mathrm{log}\frac{\mathrm{exp}(\mathrm{sim}(\mathrm{\mathbf{s}}_{i},\mathrm{\mathbf{v}}_{i})/\tau )}{\sum \nolimits_{j=1}^{\mathrm{B}} \mathrm{exp}(\mathrm{sim}(\mathrm{\mathbf{s}}_{i},\mathrm{\mathbf{v}}_{j})/\tau)},
\end{equation}
where $\mathrm{sim}(\cdot )$ denotes the cosine similarity, and $\tau$ is the learnable temperature parameter. The alignment process between these two spaces can be denoted as $\mathcal{L}_{\mathrm{InfoNCE}}^{sv}=\mathcal{L}_{s2v} + \mathcal{L}_{v2s}$, and the definition of $\mathcal{L}_{v2s}$ is similar to the $\mathcal{L}_{s2v}$. For convenience, we simplify and re-write the loss in the matrix as:
\begin{equation}
    \mathcal{L}_{\mathrm{InfoNCE}}^{sv}=\mathcal{H}(\mathrm{\mathbf{I_{B}}},\rho (\mathrm{\mathbf{SV}^{T}}))+\mathcal{H}(\mathrm{\mathbf{I_{B}}},\rho (\mathrm{\mathbf{VS}^{T}})),
\end{equation}
\begin{equation}
    \mathcal{L}_{\mathrm{InfoNCE}}^{sl}=\mathcal{H}(\mathrm{\mathbf{I_{B}}},\rho (\mathrm{\mathbf{SL}^{T}}))+\mathcal{H}(\mathrm{\mathbf{I_{B}}},\rho (\mathrm{\mathbf{LS}^{T}})),
\end{equation}
where $\mathcal{H}$ is the cross-entropy function, and $\rho$ is the standard softmax function. The $\mathrm{\mathbf{I_{B}}}$ is the identity matrix, and $\mathrm{\mathbf{S}},\mathrm{\mathbf{V}},\mathrm{\mathbf{L}}\in \mathbb{R}^{\mathrm{B}\times d}$ are the matrices that contain a batch of skeleton, vision and language embeddings. Besides, we align the skeleton and language space similarly by minimizing the $\mathcal{L}_{\mathrm{InfoNCE}}^{sl}$.

\begin{figure}[t]
\centering
\includegraphics[width=\linewidth, height=0.35\linewidth]{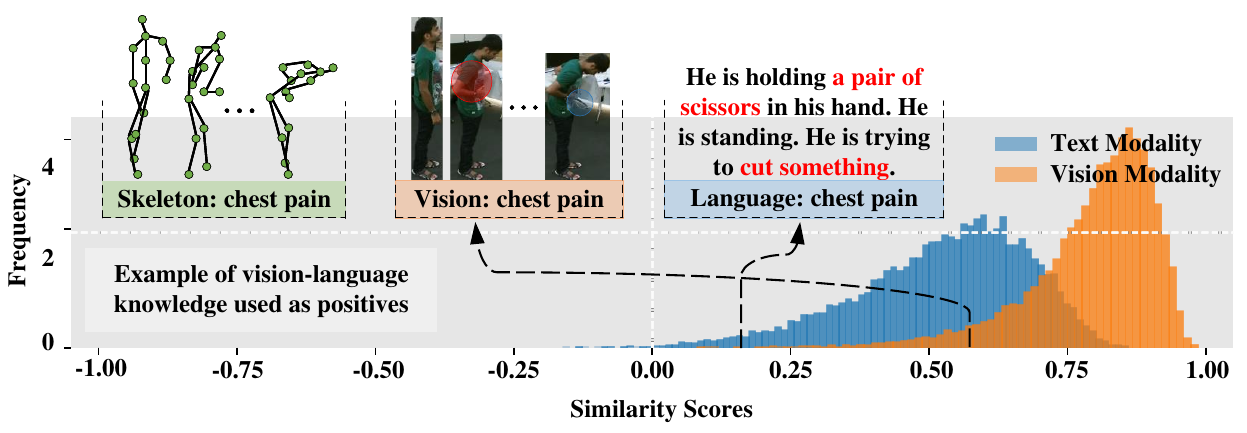}
\caption{Histogram of similarity scores for positive pairs between skeleton and vision-language knowledge representations in cross-modal space after training with original InfoNCE loss. }
\label{fig: positive_similarity}
\end{figure}

\begin{figure}[t]
\centering
\includegraphics[width=\linewidth, height=0.35\linewidth]{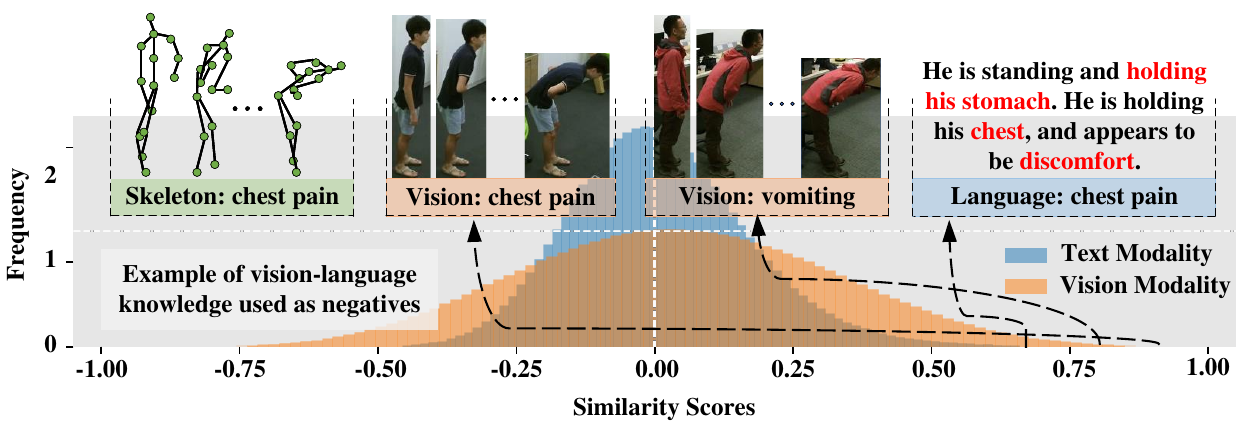}
\caption{Histogram of similarity scores for negative pairs between skeleton and vision-language knowledge representations in cross-modal space after training with original InfoNCE loss. The red ellipse highlights the movement of arms. }
\label{fig: negative_similarity}
\end{figure}

In practice, the strict assumption in InfoNCE loss to bring $(s_{i},v_{i},l_{i})$ closer and push $(s_{i},v_{j},l_{k})$ away is unreasonable for two reasons. First, as shown in Fig. \ref{fig: positive_similarity}, the ground truth vision-language knowledge prompts $(v_{i},l_{i})$ of the given skeleton $s_{i}$ may be incorrect because of hallucination in LMM and poor pose estimation, showing low semantic similarity scores among positive pairs. Second, as shown in Fig. \ref{fig: negative_similarity}, the given skeleton $s_{i}$ would be aligned to unpaired vision-language knowledge prompts $(v_{j},l_{k})$ with shared semantics to different degrees within a larger batch size, resulting in high similarity semantic scores among negative pairs. For these, the three modality pairs exhibit many-to-many correspondences among different samples in a batch. The hard alignment between these spaces based on InfoNCE loss will result in poor skeleton representations, as shown in Fig. \ref{fig:text_infonce} and Fig. \ref{fig:vision_infonce}.

\subsection{Cross-modal Alignment with Soft Targets}
\label{sec: self-distillation}
To address the limitations of using standard InfoNCE loss to learn skeleton representation on noisy joint action concept space, we attempt to design two soft targets to guide the learning of cross-modal alignment. These soft targets play crucial roles in repairing a sample with more semantically similar correspondences in other modalities, which relaxes the explicit one-to-one correspondences and improves the implicit many-to-many relationships.

\subsubsection{Intra-modal Self-similarity Guidance}
\label{sec: intra-similarity}
To estimate the many-to-many relationships within $(s_{i},v_{j},l_{k})$, we propose to utilize the intra-modal self-similarity to guide the learning of skeleton representation from the vision-language knowledge prompts. Specifically, we calculate the skeleton-to-skeleton similarity $\mathrm{\mathbf{SS}^{T}}$, vision-to-vision similarity $\mathrm{\mathbf{VV}^{T}}$, and language-to-language similarity $\mathrm{\mathbf{LL}^{T}}$ in their respective modality space as soft labels. For the training stability, we use the weighted average strategy \cite{gao2024softclip} to fuse the hard and soft labels as the softened targets to supervise the skeleton-to-vision, vision-to-skeleton, skeleton-to-language, and language-to-skeleton correspondences, respectively. These intra-modal self-similarity softened targets within a batch are defined as follows:
\begin{equation}
    \mathbf{P}^{v2s}=\beta \rho (\mathrm{\mathbf{VV}^{T}})+(1-\beta )\mathrm{\mathbf{I_{B}}},
\end{equation}
\begin{equation}
    \mathbf{P}^{l2s}=\beta \rho (\mathrm{\mathbf{LL}^{T}})+(1-\beta )\mathrm{\mathbf{I_{B}}},
\end{equation}
\begin{equation}
    \mathbf{P}^{s2v}, \mathbf{P}^{s2l}=\beta \rho (\mathrm{\mathbf{SS}^{T}})+(1-\beta )\mathrm{\mathbf{I_{B}}},
\end{equation}
where $\beta$ is a trade-off hyper-parameter set to 0.2. After replacing the hard targets $\mathrm{\mathbf{I_{B}}}$ with softened targets $\mathbf{P}_{\mathrm{intra}}^{sv}$, $\mathbf{P}_{\mathrm{intra}}^{vs}$, $\mathbf{P}_{\mathrm{intra}}^{sl}$ and $\mathbf{P}_{\mathrm{intra}}^{ls}$, the model can control the degree of attractive and repulsive forces between the skeleton and vision-language embeddings based on their intra-modal self-similarity for more generalized skeleton representation.

\subsubsection{Inter-modal Cross-consistency Boosting}
\label{sec: inter-consistency}
Although the intra-modal self-similarity relaxes the one-to-one constraint and pursues the many-to-many correspondences among the three modalities, the confidence value of the original positive pairs is still at the domination position. Especially for those faulty positive pairs, the higher confidence may restrain and submerge the process of repairing samples with more semantically similar correspondences. To mitigate this issue, we adopt the cycle consistent prediction strategy \cite{morgado2021robust} to generate the inter-nodal cross-consistency soft targets as defined as follows:
\begin{equation}
    \mathbf{Q}^{v2s}=\rho (\mathrm{\mathbf{SV}^{T}}+\mathrm{\mathbf{I_{B}}}\mathrm{\mathbf{VS}^{T}}\mathrm{\mathbf{J}}+\mathrm{\mathbf{J}}\mathrm{\mathbf{S}^{T}}\mathrm{\mathbf{V}}\mathrm{\mathbf{I_{B}}}),
    \label{equ:p_inter_vs}
\end{equation}
\begin{equation}
    \mathbf{Q}^{s2v}=\rho (\mathrm{\mathbf{VS}^{T}}+\mathrm{\mathbf{I_{B}}}\mathrm{\mathbf{SV}^{T}}\mathrm{\mathbf{J}}+\mathrm{\mathbf{J}}\mathrm{\mathbf{V}^{T}}\mathrm{\mathbf{S}}\mathrm{\mathbf{I_{B}}}),
    \label{equ:p_inter_sv}
\end{equation}
\begin{equation}
    \mathbf{Q}^{l2s}=\rho (\mathrm{\mathbf{SL}^{T}}+\mathrm{\mathbf{I_{B}}}\mathrm{\mathbf{LS}^{T}}\mathrm{\mathbf{J}}+\mathrm{\mathbf{J}}\mathrm{\mathbf{S}^{T}}\mathrm{\mathbf{L}}\mathrm{\mathbf{I_{B}}}),
    \label{equ:p_inter_ls}
\end{equation}
\begin{equation}
    \mathbf{Q}^{s2l}=\rho (\mathrm{\mathbf{LS}^{T}}+\mathrm{\mathbf{I_{B}}}\mathrm{\mathbf{SL}^{T}}\mathrm{\mathbf{J}}+\mathrm{\mathbf{J}}\mathrm{\mathbf{L}^{T}}\mathrm{\mathbf{S}}\mathrm{\mathbf{I_{B}}}),
    \label{equ:p_inter_sl}
\end{equation}
where $\mathrm{\mathbf{J}}$ is an all-one matrix. The first term of Equation (\ref{equ:p_inter_vs})-(\ref{equ:p_inter_sl}) aggregates the information from its cross opposite modality for dual-direction alignment, \textit{i.e.}, alignment from vision $\mathrm{\mathbf{v}}_{i}$ to skeleton $\mathrm{\mathbf{s}}_{j}$ is based on the probability that skeleton $\mathrm{\mathbf{s}}_{i}$ matched with vision $\mathrm{\mathbf{v}}_{j}$. Moreover, the second and third terms are designed to avoid positive pairs that are not sound correspondences for boosting negative pairs with high semantic similarity in cross-modalities.

\subsection{Progressive Self-distillation Training Objectives}
\label{sec: progressive}
Here, we introduce a progressive self-knowledge distillation framework\cite{kim2021self} where the two soft targets are dynamically updated at each epoch by the teacher network to guide the learning of the student network. Compared to the conventional knowledge distillation methods utilized the static and separate pre-trained teacher network, the main difference of this self-distillation strategy is that it regards the same model architecture as the student and teacher, which has the advantage of reducing computation and memory costs. Meanwhile, as the training goes on, the well-trained teacher can provide more confidence in learning soft labels for students, which mitigates the adverse effects of noisy pairs and enhances the generalization capability.

To align skeleton space with vision-knowledge spaces, we adopt a progressive scheme \cite{andonian2022robust} to gradually control and balance the roles of two soft targets throughout training. Specifically, we partition a batch of pairs into two subsets with the refreshed hyper-parameter $\alpha \in [0,1]$ at each epoch. The intra-modal self-similarity soft target is utilized to repair the first subset data $\mathrm{B_{intra}}=\left \lfloor \alpha \mathrm{B} \right \rfloor$ for estimation of the many-to-many relationships, while the inter-class modal cross-consistency is employed to guide the second subset $\mathrm{B_{inter}}=\mathrm{B}-\left \lfloor \alpha \mathrm{B} \right \rfloor$ for boosting negative pairs with high semantic similarity. At the initial training phase, the value of $\alpha$ should be large for training stability. With the increased training iterations, we gradually decrease the value of $\alpha$ by cosine annealing schedule \cite{li2020few} to increase the contribution of alignment extent between skeleton and high semantic similarity vision-language knowledge prompts. The start value is 0.9, and the end value is 0.1. In this way, the skeleton representation can be better learned in the noisy cross-modal pairs.

Altogether, the final training objective can be formulated as follows:
\begin{equation}
    \mathcal{L}_{\mathrm{Soft}}=\mathcal{L}_{\mathrm{Soft}}^{sv}+\mathcal{L}_{\mathrm{Soft}}^{sl},
\end{equation}
\begin{equation}
\begin{split}
    \mathcal{L}_{\mathrm{Soft}}^{sv}=\alpha&[\mathcal{H}(\mathbf{P}_{\mathrm{intra}}^{s2v},\rho (\mathrm{\mathbf{\overline{S}}\mathbf{V}^{T}}))+\mathcal{H}(\mathbf{P}_{\mathrm{intra}}^{v2s}, \rho (\mathrm{\mathbf{\overline{V}}\mathbf{S}^{T}})]+\\
(1-\alpha)&[\mathcal{H}(\mathbf{Q}_{\mathrm{inter}}^{s2v},\rho (\mathrm{\mathbf{\underline{S}}\mathbf{V}^{T}}))+\mathcal{H}(\mathbf{Q}_{\mathrm{inter}}^{v2s}, \rho (\mathrm{\mathbf{\underline{V}}\mathbf{S}^{T}})],
\end{split}
\end{equation}
\begin{equation}
\begin{split}
    \mathcal{L}_{\mathrm{Soft}}^{sl}=\alpha&[\mathcal{H}(\mathbf{P}_{\mathrm{intra}}^{s2l},\rho (\mathrm{\mathbf{\overline{S}}\mathbf{L}^{T}}))+\mathcal{H}(\mathbf{P}_{\mathrm{intra}}^{l2s}, \rho (\mathrm{\mathbf{\overline{L}}\mathbf{S}^{T}})]+\\
(1-\alpha)&[\mathcal{H}(\mathbf{Q}_{\mathrm{inter}}^{s2l},\rho (\mathrm{\mathbf{\underline{S}}\mathbf{L}^{T}}))+\mathcal{H}(\mathbf{Q}_{\mathrm{inter}}^{l2s}, \rho (\mathrm{\mathbf{\underline{L}}\mathbf{S}^{T}})],
\end{split}
\end{equation}
where $\mathbf{P}_{\mathrm{intra}}^{*}$ is the first $\mathrm{B_{intra}}$ rows of intra-modal self-similarity soft targets, and the $\mathbf{Q}_{\mathrm{inter}}^{*}$ is the last $\mathrm{B_{inter}}$ rows of inter-modal cross-consistency soft targets. $\overline{S}, \overline{V}, \overline{L} \in \mathbb{R}^{\mathrm{B_{intra}} \times d}$ denotes the first $\mathrm{B_{intra}}$ rows of skeleton, vision, and language embeddings, respectively. Meanwhile, $\underline{S}, \underline{V}, \underline{L} \in \mathbb{R}^{\mathrm{B_{inter}} \times d}$ denotes the last $\mathrm{B_{inter}}$ rows of respective embeddings. 

In the inference phase, only the skeleton and encoder $E_{s}$ are used for skeleton-based action recognition. The vision-language prompts and respective encoders will be dropped.

\section{Experiments}
\label{sec:experiments}

\subsection{Datasets}
\label{ex:datasets}

\subsubsection{NTU RGB+D 60 Dataset \cite{shahroudy2016ntu}}
\label{ex:ntu_rgb_d_60}
It comprises 60 predefined action categories and 56,880 action samples, containing multiple modalities, including RGB, depth map, skeleton, and infrared (IR) video. The skeleton sequences contain a maximum of two human skeletons, represented by the 3D coordinates of 25 joints. The RGB video has a resolution of $1920\times 1080$. Two benchmarks are provided in this dataset: cross-subject (Xsub) and cross-view (Xview). In Xsub, 20 subjects are used for training, while the remaining subjects are used for testing. In Xview, sequences from views 2 and 3 are designated for training, and the sequences from view 1 are used for testing.    

\subsubsection{NTU RGB+D 120 Dataset \cite{liu2019ntu}}
\label{ex:ntu_rgb_d_120}
It is an extension of NTU RGB+D 60 with 120 predefined action categories and 114,480 action samples, which have the same modality as NTU RGB+D 60. Similarly, this dataset provides two benchmarks: cross-subject (Xsub) and cross-setup (Xset). In Xsub, sequences from 53 subjects are utilized for training, while the remaining subjects are reserved for testing. In Xset, sequences from even camera IDs are designated for training, and the rest are employed for testing.

\subsubsection{PKU-MMD Dataset \cite{liu2017pku}}
\label{ex:pku_mmd}
It comprises 51 action categories and almost 20000 action samples with the same multi-modalities as the NTU series datasets. The dataset also provides two benchmarks: cross-subject (Xsub) and cross-view (Xview). In Xsub, sequences from 57 subjects belong to the training set and 9 subjects for the testing. In Xview, sequences from the middle and right views are utilized for training, and the left is used for testing. In this experiment, we adopt the cross-subject benchmark following the previous studies.

\begin{table*}
  \caption{Comparison to the state-of-the-art methods for action recognition accuracy on the NTU RGB+D 60, NTU RGB+D 120, and PKU-MMD II datasets under the linear evaluation protocol. The best and second-best results are highlighted in \textbf{bold} and \underline{underlined}, respectively. The J refers to the joint modality, the M denotes the motion modality, and the B is the bone modality.}
  \centering
  \begin{tabular}{l l c c c c c c c}
    \toprule
    \multirow{2.5}{*}{Method} & \multirow{2.5}{*}{Publication} & \multirow{2.5}{*}{Architecture} & \multirow{2.5}{*}{Modality} & \multicolumn{2}{c}{NTU60} & \multicolumn{2}{c}{NTU120} & PKU-MMD II \\
    \cmidrule(r){5-6} \cmidrule(r){7-8} \cmidrule(r){9-9}
     & & & & Xsub & Xview & Xsub & Xset & Xsub \\
    \midrule
    \rowcolor{gray!10} \multicolumn{9}{l}{\textit{Single-stream}:} \\
    AimCLR \cite{guo2022contrastive} & AAAI'22 & GCN & J & 74.3 & 79.7 & 63.4 & 63.4 & - \\
    CPM \cite{zhang2022contrastive} & ECCV'22 & GCN & J & 78.7 & 84.9 & 68.7 & 69.6 & 48.3 \\
    CMD \cite{mao2022cmd} & ECCV'22 & GRU & J & 79.8 & 86.9 & 70.3 & 71.5 & 43.0 \\
    HaLP \cite{shah2023halp} & CVPR'23 & GRU & J & 79.7 & 86.8 & 71.1 & 72.2 & 43.5 \\
    ActCLR \cite{lin2023actionlet} & CVPR'23 & GCN & J & 80.9 & 86.7 & 69.0 & 70.5 & - \\
    DMMG \cite{guan2023dmmg} & TIP'23 & GCN & J & 82.1 & 87.1 & 69.6 & 70.1 & - \\
    CSTCN \cite{wang2023learning} & TMM'23 & GRU & J & 83.1 & 88.7 & 72.5 & 77.4 & 48.0 \\
    UmURL \cite{sun2023unified} & ACM MM'23 & Transformer & J & 82.3 &
    \underline{89.8} & 73.5 & 74.3 & \underline{52.1} \\
    PCM$^{3}$ \cite{PCM_zhang2023} & ACM MM'23 & GRU & J & \underline{83.9} & \textbf{90.4} & \textbf{76.5} & \underline{77.5} & 51.5 \\
    KTCL \cite{wang2024localized} & TMM'24 & Transformer & J & 82.4 & 89.4 & 74.4 & 74.5 & - \\
    \rowcolor{yellow!10} \textbf{C$^2$VL (Ours)} & This work & GCN & J & \textbf{84.4} & \underline{89.8} & \underline{76.0} & \textbf{78.7} & \textbf{52.6}\\
    \midrule
    \rowcolor{gray!10} \multicolumn{9}{l}{\textit{Multi-stream:}}\\
    3s-CrosSCLR \cite{li20213d} & CVPR'21 & GCN & J+M+B & 77.8 & 83.4 & 67.9 & 66.7 & 21.2 \\
    3s-AimCLR \cite{guo2022contrastive} & AAAI'22 & GCN & J+M+B & 78.9 & 83.8 & 68.2 & 68.8 & 39.5 \\
    3s-CMD \cite{mao2022cmd} & ECCV'22 & GRU & J+M+B & 84.1 & 90.9 & 74.7 & 76.1 & 52.6 \\
    3s-CPM \cite{zhang2022contrastive} &  ECCV'22 & GCN & J+M+B & 83.2 & 87.0 & 73.0 & 74.0 & 51.5 \\
    3s-ActCLR \cite{lin2023actionlet} & CVPR'23 & GCN & J+M+B & 84.2 & 88.8 & 74.3 & 75.7 & - \\
    2s-DMMG \cite{guan2023dmmg} & TIP'23 & GCN & J+M & 84.2 & 89.3 & 72.7 & 72.4 & - \\
    3s-CSTCN \cite{wang2023learning} & TMM'23 & GRU & J+M+B & 85.8 & 92.0 & 77.5 & 78.5 & 53.9 \\
    3s-UmURL \cite{sun2023unified} & ACM MM'23 & Transformer & J+M+B & 84.4 & 91.4 & 75.9 & 77.2 & 54.3 \\
    3s-PCM$^{3}$ \cite{PCM_zhang2023} & ACM MM'23 & GRU & J+M+B & \underline{87.4} & \textbf{93.1} & \underline{80.0} & \underline{81.2} & \underline{58.2} \\
    \rowcolor{yellow!10} \textbf{3s-C$^2$VL (Ours)} & This work & GCN & J+M+B & \textbf{88.3} & \underline{92.8} & \textbf{82.5} & \textbf{84.3} & \textbf{60.0} \\
    \bottomrule
  \end{tabular}
  \label{tab:linear}
\end{table*}

\subsection{Implementation Details}
\label{sec:implementation_details}
All experiments are conducted using the PyTorch framework on an NVIDIA A100 GPU. Each sample is down-sampled to 64 frames using the data pre-processing code from \cite{xiang2023generative}. We use ST-GCN \cite{yan2018spatial} as our skeleton GCN encoder. For the image-text encoder, we utilize the pre-trained \textbf{ViT-L/14@336px} model from CLIP \cite{radford2021learning} and freeze its parameters during training. The temperature parameter $\tau$ is 0.07. In the pre-training phase, we train the model with the SGD optimizer for 150 epochs with a batch size 400. The initial learning rate is 0.1 and reduced by a factor of 0.1 at epochs 130 and 140. Weight decay is set to 5e-4 following the strategy in \cite{xiang2023generative}. In the inference stage, we utilize multiple evaluation protocols, including the linear evaluation protocol, finetune protocol, semi-supervised evaluation protocol(1\%, 5\% or 10\% labeled data), KNN evaluation protocol, and transfer learning evaluation protocol for action recognition.
Meanwhile, following the previous studies \cite{guo2022contrastive, zhang2022contrastive, mao2022cmd, shah2023halp, lin2023actionlet, guan2023dmmg, wang2023learning, sun2023unified, PCM_zhang2023, wang2024localized}, several tasks (cross-view, cross-subject, and cross-setup) with different official training-testing split strategies are used to comprehensively validate the algorithm, preventing overfitting in the testing set.

\subsection{Comparison to the State-of-the-art}
\label{sec:performance_comparison}
To comprehensively evaluate the performance of our method, we conduct comparisons with other state-of-the-art methods across various settings.

\subsubsection{Linear Evaluation}
\label{ex:comparision_linear}
The linear evaluation protocol applies a fully connected layer after the parameter-fixed ST-GCN to classify. As shown in Table \ref{tab:linear}, we found that both single-stream and three-stream performance can achieve the best or the second-best results on NTU RGB+D 60 \& 120 and PKU-MMD II datasets, indicating that vision-language knowledge prompts have a solid capacity to guide GCN for generalization representation.

\subsubsection{Finetune Evaluation}
\label{ex:comparision_finetune}
We pre-train ST-GCN and finetune the entire network for action recognition, applying a fully connected layer. Table \ref{tab:finetune} compares the results obtained through our pre-training and other training paradigms. Notably, our training paradigm performs better in capturing action patterns and details from vision-language knowledge prompts, outperforming some supervised learning methods (Shift-GCN) and even surpassing state-of-the-art self-supervised methods.

\begin{table}
    \renewcommand{\arraystretch}{1.2}
  \caption{Comparison of action recognition results employing different training paradigms (supervised and self-supervised learning) on the NTU dataset. Supervised learning utilizes one-hot labels for training models, while self-supervised learning utilizes the pretext task to pre-train models and fine-tune them with labels. The best and second-best results are highlighted in \textbf{bold} and \underline{underlined}, respectively. SL: Supervised Learning, SSL: Self-supervised Learning.}
  \centering
  \begin{tabular}{l c c c c c}
    \toprule
    \multirow{2.5}{*}{Method} & \multirow{2.5}{*}{Paradigm} & \multicolumn{2}{c}{NTU60} & \multicolumn{2}{c}{NTU120} \\
    \cmidrule(r){3-4} \cmidrule(r){5-6}
     & & Xsub & Xview & Xsub & Xset \\
    \midrule
    \rowcolor{gray!10} \multicolumn{6}{l}{\textit{Single-stream}:} \\
    ST-GCN \cite{yan2018spatial} & SL & 81.5 & 88.3 & 70.7 & 73.2 \\
    Shift-GCN \cite{cheng2020skeleton} & SL & 87.8 & 95.1 & 80.9 & 83.2 \\
    \hdashline
    SkeletonCLR \cite{li20213d} & SSL & 82.2 & 88.9 & 73.6 & 75.3 \\
    AimCLR \cite{guo2022contrastive} & SSL & 83.0 & 89.2 & 77.2 & 76.1 \\
    CPM \cite{zhang2022contrastive} & SSL & 84.8 & 91.1 & 78.4 & 78.9 \\
    ActCLR \cite{lin2023actionlet} & SSL & \underline{85.8} & \underline{91.2} & \underline{79.4} & \underline{80.9} \\
    \rowcolor{yellow!10} \textbf{C$^2$VL (Ours)} & SSL & \textbf{88.9} & \textbf{93.2} & \textbf{83.4} & \textbf{84.9} \\
    \midrule
    \rowcolor{gray!10} \multicolumn{6}{l}{\textit{Multi-stream:}}\\
    3s-ST-GCN \cite{yan2018spatial} & SL & 85.2 & 91.4 & 77.2 & 77.1 \\
    4s-Shift-GCN \cite{cheng2020skeleton} & SL & 90.7 & 96.5 & 85.9 & 87.6 \\ 
    \hdashline
    3s-CrosSCLR \cite{li20213d} & SSL & 86.2 & 92.5 & 80.5 & 80.4 \\
    3s-AimCLR \cite{guo2022contrastive} & SSL & 86.9 & 92.8 & 80.1 & 80.9 \\
    3s-ActCLR \cite{lin2023actionlet} & SSL & \underline{88.2} & \underline{93.9} & \underline{82.1} & \underline{84.6} \\
    \rowcolor{yellow!10} \textbf{3s-C$^2$VL (Ours)} & SSL & \textbf{91.8} & \textbf{95.7} & \textbf{87.8} & \textbf{89.2}\\
    \bottomrule
  \end{tabular}
  \label{tab:finetune}
\end{table}

\subsubsection{Semi-supervised Evaluation}
\label{ex:comparision_semi}
Our approach involves pre-training with the entire training data and finetuning the classifier with only 1\%, 5\%, and 10\% of the labeled data, respectively. Table \ref{tab:semi} demonstrates that our method significantly renews the accuracy of the state-of-the-art methods, indicating its effectiveness and strong generalization capabilities in learning high-quality task-agnostic representations without relying on extensive labels.

\begin{table}
  \caption{Comparison of action recognition results under semi-supervised evaluation protocol on NTU RGB+D 60 dataset. The best and second-best results are highlighted in \textbf{bold} and \underline{underlined}, respectively.}
  \centering
  \begin{tabular}{lcccccc}
    \toprule
    \multirow{2.5}{*}{Method} & \multicolumn{3}{c}{Xsub} & \multicolumn{3}{c}{Xview} \\
    \cmidrule(r){2-4} \cmidrule{5-7}
     & 1\% & 5\% & 10\% & 1\% & 5\% & 10\%  \\
    \midrule
    ASSL \cite{si2020adversarial} & - & 57.3 & 64.3 & - & 63.6 & 69.8 \\
    MS$^{2}$L \cite{lin2020ms2l} & 33.1 & - & 65.1 & - & - & - \\
    MCC \cite{su2021self} & - & 47.4 & 60.8 & - & 53.3 & 65.8 \\
    ISC \cite{thoker2021skeleton} & 35.7 & 59.6 & 65.9 & 38.1 & 65.7 & 72.5 \\
    Colorization \cite{yang2021skeleton} & 48.3 & 65.7 & 71.7 & 52.5 & 70.3 & 78.9 \\
    Hi-TRS \cite{chen2022hierarchically} & 39.1 & 63.3 & 70.7 & 42.9 & 68.3 & 74.8 \\
    CPM \cite{zhang2022contrastive} & 56.7 & - & 73.0 & 57.5 & - & 77.1 \\
    CMD \cite{mao2022cmd} & 50.6 & 71.0 & 75.4 & 53.0 & 75.3 & 80.2 \\
    HiCo \cite{dong2023hierarchical} & 54.4 & - & 73.0 & 54.8 & - & 78.3 \\
    PCM$^{3}$ \cite{PCM_zhang2023} & 53.8 & - & \underline{77.1} & 53.1 & - & \underline{82.8} \\
    UmURL \cite{sun2023unified} & \underline{58.1} & \underline{72.5} & - & \underline{58.3} & \underline{76.8} & - \\
    \rowcolor{yellow!10} \textbf{C$^2$VL (Ours)} & \textbf{69.3} & \textbf{79.4} & \textbf{81.8} & \textbf{69.1} & \textbf{82.1} & \textbf{85.5} \\
    \bottomrule
  \end{tabular}
  \label{tab:semi}
\end{table}

\subsubsection{KNN Evaluation}
\label{ex:comparision_knn}
In this experiment, we utilize the K-nearest neighbors (KNN) as a classifier for action retrieval based on the obtained skeleton representation from the pre-training model. The results on NTU60 and NTU120 datasets are shown in Table \ref{tab:knn}. Our method achieves the best results among all tasks, surpassing the previous methods by a large margin. Moreover, it demonstrates that the obtained skeleton representation is more discriminative than other methods.

\begin{table}
  \caption{Comparison to the state-of-the-art methods for action retrieval with the joint stream on NTU RGB+D 60 and NTU RGB+D 120 datasets. The best and second-best results are highlighted in \textbf{bold} and \underline{underlined}, respectively.}
  \centering
  \begin{tabular}{lcccc}
    \toprule
    \multirow{2.5}{*}{Method} & \multicolumn{2}{c}{NTU60} & \multicolumn{2}{c}{NTU120} \\
    \cmidrule(r){2-3} \cmidrule(r){4-5}
     & Xsub & Xview & Xsub & Xset \\
    \midrule
    LongT GAN \cite{zheng2018unsupervised} & 39.1 & 48.1 & 31.5 & 35.5 \\
    P\&C \cite{su2020predict} & 50.7 & 76.3 & 39.5 & 41.8 \\
    ISC \cite{thoker2021skeleton} & 62.5 & 82.6 & 50.6 & 52.3 \\
    AimCLR \cite{guo2022contrastive} & 62.0 & 71.5 & - & - \\
    CMD \cite{mao2022cmd} & 70.6 & 85.4 & 58.3 & 60.9 \\
    HiCLR \cite{zhang2023hierarchical} & 67.3 & 75.3 & - & - \\
    HiCo \cite{dong2023hierarchical} & 68.3 & 84.8 & 56.6 & 59.1 \\
    HaLP \cite{shah2023halp} & 65.8 & 83.6 & 55.8 & 59.0 \\
    UmURL \cite{sun2023unified} & 71.3 & 88.3 & 58.5 & 60.9 \\
    PCM$^{3}$ \cite{PCM_zhang2023} & \underline{73.7} & \underline{88.8} & \underline{63.1} & \underline{66.8} \\
    \rowcolor{yellow!10} \textbf{C$^2$VL (Ours)} & \textbf{78.0} & \textbf{88.8} & \textbf{64.0} & \textbf{68.8} \\
    \bottomrule
  \end{tabular}
  \label{tab:knn}
\end{table}

\subsubsection{Transfer Learning Evaluation}
\label{ex:comparision_transfer}
We evaluate the transfer generalizability of the obtained skeleton representation in this part. Specifically, we use the model pre-trained on the NTU60 and NTU120 datasets to finetune the PKU-MMD II dataset. As shown in Table \ref{tab:transfer}, our method outperforms the competitors by a large margin, showing the excellent transferability of our learned skeleton representation, which has enormous potential in data-scarcity scenarios.

\begin{table}
  \caption{Comparison to the state-of-the-art methods with transfer learning on Xsub task. The source dataset is NTU RGB+D 60 or NTU RGB+D 120, and the target dataset is PKU-MMD II. The best and second-best results are highlighted in \textbf{bold} and \underline{underlined}, respectively.}
  \centering
  \begin{tabular}{l c c}
    \toprule
    \multirow{2.5}{*}{Method} & \multicolumn{2}{c}{Transfer to PKU-MMD II} \\
    \cmidrule(r){2-3} 
     & NTU60 & NTU120 \\
    \midrule
    LongT GAN \cite{zheng2018unsupervised} & 44.8 & - \\
    M$^2$L \cite{lin2020ms2l} & 45.8 & - \\
    ISC \cite{thoker2021skeleton} & 45.9 & - \\
    CrosSCLR \cite{li20213d} & 54.0 & 52.8 \\
    CMD \cite{mao2022cmd} & 56.0 & 57.0 \\
    HiCo \cite{dong2023hierarchical} & 56.3 & 55.4 \\
    UmURL \cite{sun2023unified} & \underline{58.2} & \underline{57.6} \\
    \rowcolor{yellow!10} \textbf{C$^2$VL (Ours)} & \textbf{67.5} & \textbf{69.6} \\
    \bottomrule
  \end{tabular}
  \label{tab:transfer}
\end{table}

\subsection{Evaluation of Knowledge Necessity}
\label{ex:analysis_multi}
To further explore the significance of distinct modal information during the pre-training phase, we employed single-modal knowledge prompts as the supervision for pre-training. Table \ref{tab:analysis_modality} presents the action recognition accuracy of linear evaluation protocol across different modalities. Notably, vision-language knowledge prompts exhibit the best performance and offer valuable guidance and assistance for skeleton representation learning. In addition, language knowledge prompts tend to offer more fine-grained and local semantic details than vision knowledge prompts cannot provide.

\begin{table}
  \caption{Analysis of different knowledge prompts on NTU RGB+D 60 Xsub dataset with the joint stream.}
  \centering
  \begin{tabular}{c c c c}
    \toprule
    \multicolumn{2}{c}{Knowledge} & \multirow{2.5}{*}{Xsub} & \multirow{2.5}{*}{Xview}\\
    \cmidrule{1-2} 
    Vision & Language & & \\
    \midrule
    \checkmark & \ding{55} & 78.6 & 81.9\\
    \ding{55} & \checkmark & 83.8 & 88.6\\
    \checkmark & \checkmark & 84.4 & 89.8\\
    \bottomrule
  \end{tabular}
  \label{tab:analysis_modality}
\end{table}

\subsection{Ablation Study}
\label{ex:ablation_study}

\subsubsection{Influences of Components}
\label{ex:components}
We evaluate the efficiency of different components during the pre-training phase in Table \ref{tab:analysis_components}. It is noted that the proposed soft targets are necessary as they can draw the unpaired data with high similarity closer, fostering more robust task-agnostic skeleton representations. Additionally, dynamic partitioning and progressive mechanisms have effectively controlled the degree of pair-bringing.

\begin{table}
    \renewcommand{\arraystretch}{1.2}
  \caption{Analysis of different components on NTU RGB+D 60 Xsub dataset with the joint stream.}
  \centering
  \begin{tabular}{c c c c c}
    \toprule
    \multicolumn{2}{c}{Soft Targets} & \multicolumn{2}{c}{Contrastive Strategies} & \multirow{3}{*}{Acc (\%)} \\
    \cmidrule(r){1-2} \cmidrule(r){3-4}
    \makecell[c]{Intra-\\Similarity} & \makecell[c]{Inter-\\Consistency} & \makecell[c]{Dynamic\\Partitioning} & \makecell[c]{Progressive\\Control} & \\ 
    \midrule
    \ding{55} & \ding{55} & \ding{55} & \ding{55} & 82.0\\
    \checkmark & \ding{55} & \ding{55} & \ding{55} & 82.8\\
    \ding{55} & \checkmark & \ding{55} & \ding{55} & 83.0\\
    \checkmark & \checkmark & \ding{55} & \ding{55} & 83.4\\
    \hdashline
    \checkmark & \checkmark & \checkmark & \ding{55} & 81.9\\
    \checkmark & \checkmark & \ding{55} & \checkmark & 84.0\\
    \checkmark & \checkmark & \checkmark & \checkmark & 84.4\\
    \bottomrule
  \end{tabular}
  \label{tab:analysis_components}
\end{table}

\subsubsection{Influences of Skeleton Encoders}
\label{ex:backbone}
Furthermore, we analyze the performance of our method across different skeleton encoders, including GRU, GCN, and Transformer architectures. As shown in Table \ref{tab:analysis_skeleton}, the GCN series encoders achieve the promised performance, which retains the topology structure of skeleton data and explores the spatial-temporal characteristic of actions. It is noted that the performance positively correlates with the ability of encoders. However, the results of UmURL drop a lot because its Transformer architecture decouples and impairs the spatial-temporal topology property of the skeleton. The PCM$^{3}$ with GRU architecture neglects the spatial structure of the skeleton, resulting in inferior modeling of spatial movements. Therefore, our method can effectively guide the learning of skeleton representation from the complete topology skeleton.

\begin{table}
  \caption{The influence of different skeleton encoders on NTU RGB+D 60 dataset with the joint stream.}
  \centering
  \begin{tabular}{lccc}
    \toprule
    \multirow{2.5}{*}{Backbone} & \multirow{2.5}{*}{Architecture} & \multicolumn{2}{c}{NTU60} \\
    \cmidrule{3-4}
     & & Xsub & Xview \\
    \midrule
    \rowcolor{yellow!10} ST-GCN (Our used)  & GCN & 84.4 & 89.8 \\
    \hdashline
    PCM$^3$ & GRU & 74.1 (\textcolor{blue}{-10.3}) & 80.1 (\textcolor{blue}{-9.7})\\
    UmURL & Transformer & 79.0 (\textcolor{blue}{-5.4}) & 84.7 (\textcolor{blue}{-5.1})\\
    MS-G3D & GCN & 84.5 (\textcolor{red}{+0.1}) &  89.9 (\textcolor{red}{+0.1})\\
    CTR-GCN & GCN & 84.5 (\textcolor{red}{+0.1}) & 90.4 (\textcolor{red}{+0.6}) \\
    \bottomrule
  \end{tabular}
  \label{tab:analysis_skeleton}
\end{table}

\subsubsection{Influences of the Hyper-parameters}
\label{ex:parameters}
\begin{figure}
\subfloat[]{
    \includegraphics[width=0.48\linewidth]{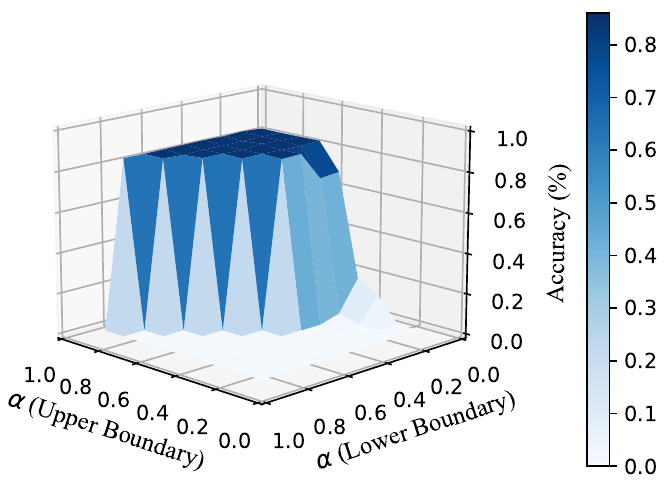}
    \label{fig:alpha_ntu60}
}
\subfloat[]{
    \includegraphics[width=0.45\linewidth]{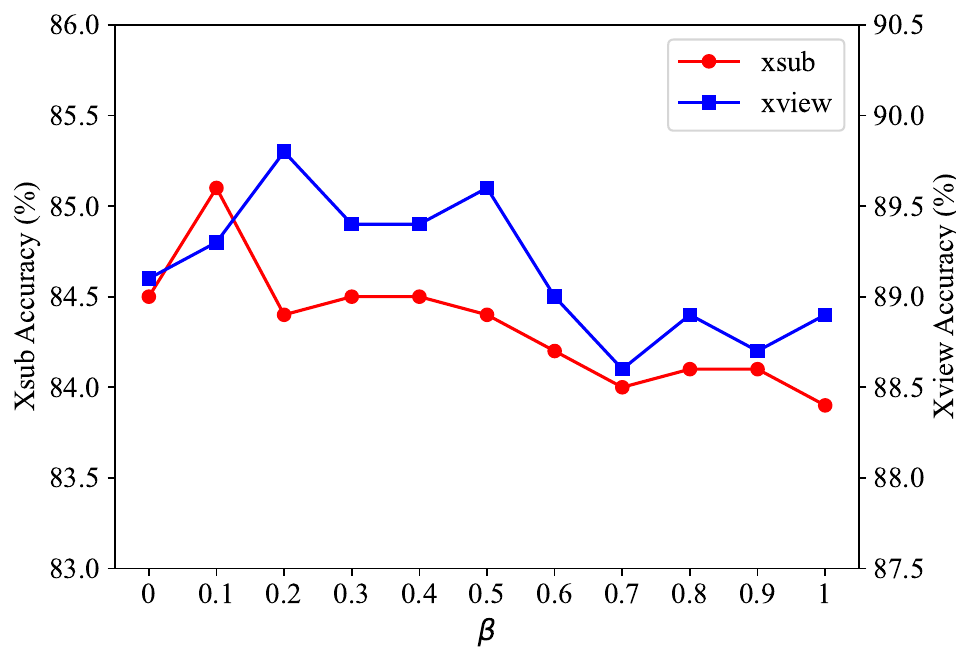}
    \label{fig:beta_ntu60}
}
\caption{(a) The influence of progressive extent $\alpha$ at different ratios on the NTU RGB+D 60 dataset. (b) The influence of intra-modal similarity guidance extent $\beta$ at different ratios on the NTU RGB+D 60 dataset.}
\label{fig:hyperparameters}
\end{figure}

As shown in Fig. \ref{fig:hyperparameters}, we evaluate the influence of the hyper-parameters. For the hyper-parameter $\alpha$, we find that the teacher network should dominate the early learning period, and the progressive extent should be large for full exploration to align unpaired cross-modal data with high similarity. Meanwhile, the hyper-parameter $\beta$ exhibits a slight increase and decrease, showing that the hard alignment and the robustness of our proposed intra-modal similarity guidance should be required.

\begin{figure}
\subfloat[Language (InfoNCE)]{
    \includegraphics[width=0.45\linewidth, height=0.3\linewidth]{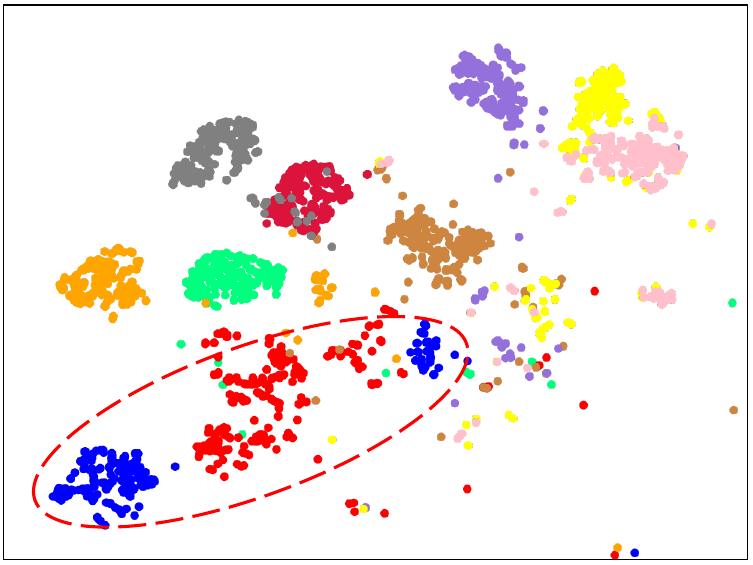}
    \label{fig:text_infonce}
}
\subfloat[Language (Ours)]{
    \includegraphics[width=0.45\linewidth, height=0.3\linewidth]{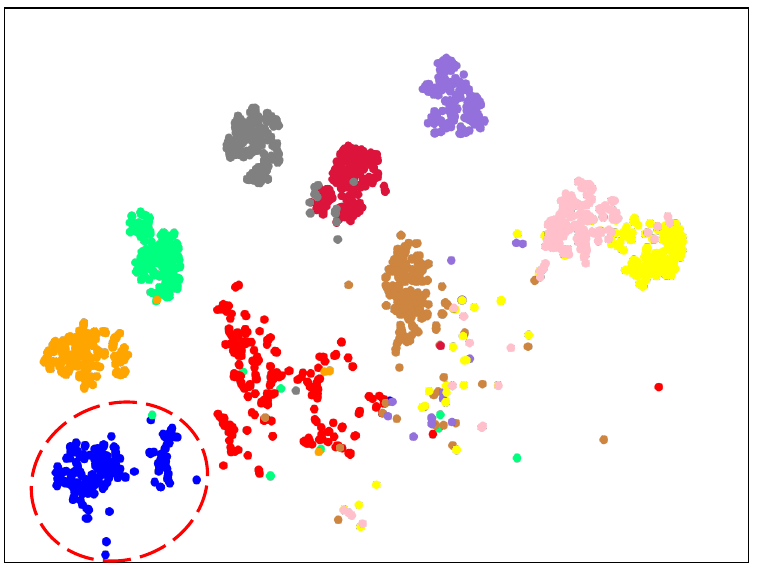}
    \label{fig:text_ours}
}
\\
\subfloat[Vision (InfoNCE)]{
    \includegraphics[width=0.45\linewidth, height=0.3\linewidth]{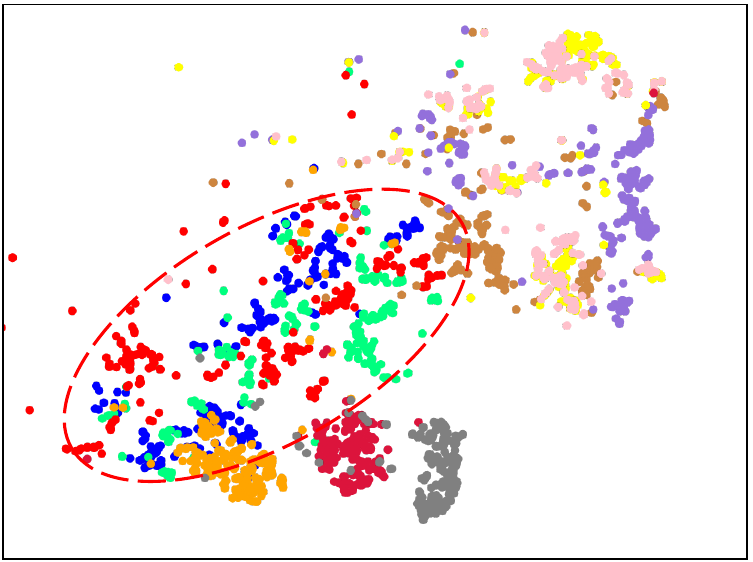}
    \label{fig:vision_infonce}
}
\subfloat[Vision (Ours)]{
    \includegraphics[width=0.45\linewidth, height=0.3\linewidth]{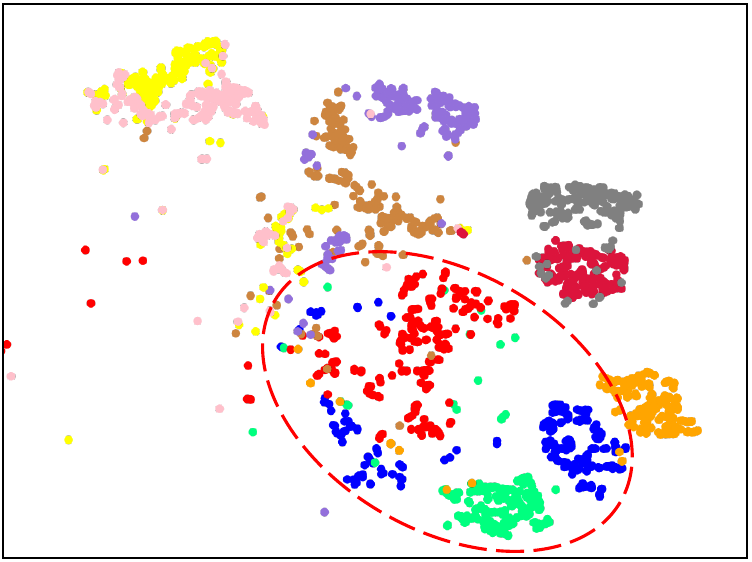}
    \label{fig:vision_ours}
}
\caption{The t-SNE visualization of action representations obtained by original InfoNCE loss (\ref{fig:text_infonce}, \ref{fig:vision_infonce}) and our proposed loss(\ref{fig:text_ours}, \ref{fig:vision_ours}) with the different modality as the supervision on the NTU RGB+D 60 Xsub dataset. 10 classes from the testing set are randomly selected for visualization, where points of the same color correspond to the same action category.}
\label{fig:tsne}
\end{figure}

\subsection{Qualitative Analysis}
\label{sec:qualitative_analysis}

\subsubsection{t-SNE Visualization of Representations}
\label{sec:tsne}
We employed t-SNE \cite{van2008visualizing} on the NTU RGB+D 60 Xsub benchmark to visualize the embedding distribution obtained by InfoNCE loss and our proposed loss with vision and language knowledge prompts as supervision, respectively. As shown in Fig. \ref{fig:tsne}, we select 10 classes for visualization, where points of the same color correspond to the same action category. It reveals that language provides better guidance for representations of discriminative skeletons than vision knowledge. Moreover, the proposed soft alignment and progressive self-distillation strategies have the advantage of clustering similar representations, resulting in better inter-class compactness and intra-class separability.

\subsubsection{Confusion Matrix}
\label{confusion}
\begin{figure*}
\subfloat[Language (InfoNCE)]{
    \includegraphics[width=0.24\linewidth]{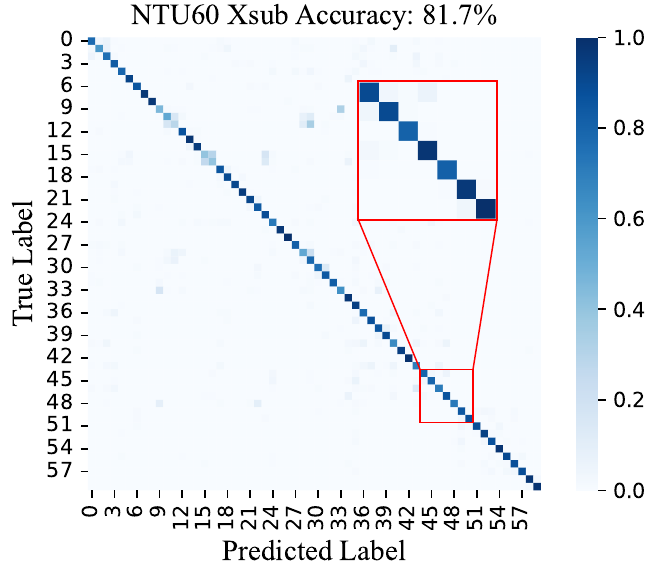}
    \label{fig:cm_text_infonce}
}
\subfloat[Language (Ours)]{
    \includegraphics[width=0.24\linewidth]{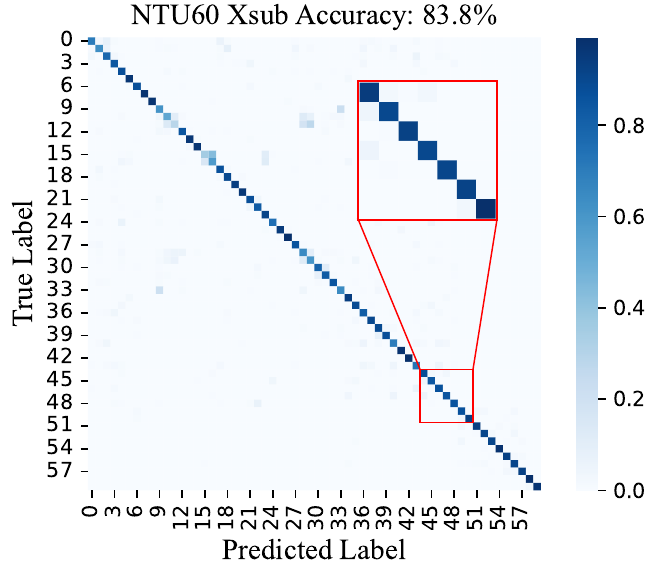}
    \label{fig:cm_text_ours}
}
\subfloat[Vision (InfoNCE)]{
    \includegraphics[width=0.24\linewidth]{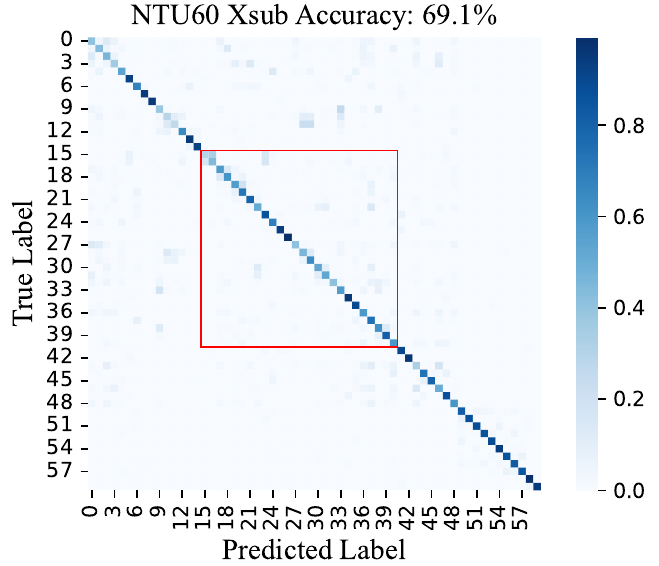}
    \label{fig:cm_vision_infonce}
}
\subfloat[Vision (Ours)]{
    \includegraphics[width=0.24\linewidth]{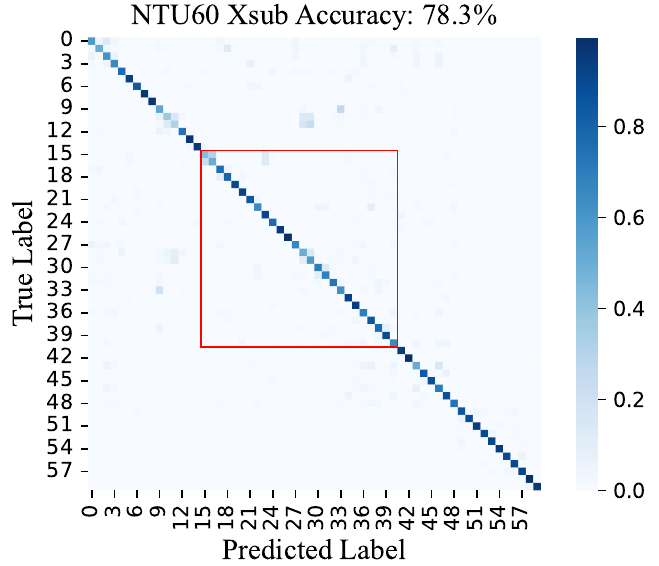}
    \label{fig:cm_vision_ours}
}
\caption{The confusion matrices of original InfoNCE loss (\ref{fig:cm_text_infonce}, \ref{fig:cm_vision_infonce}) and our proposed loss (\ref{fig:cm_text_ours}, \ref{fig:cm_vision_ours}) with the different modality as the supervision on the NTU RGB+D 60 Xsub dataset under the linear probe evaluation.}
\label{fig:cm}
\end{figure*}

Here, we draw the confusion matrices of skeleton-based action recognition results with the different modalities as the supervision based on InfoNCE loss and our proposed strategies. As shown in Fig. \ref{fig:cm}, we can find that the proposed soft alignment strategies significantly improve results, especially with the vision knowledge prompts as the supervision. Compared to the InfoNCE loss, our proposed loss improves recognition accuracy by nearly 10\% with the vision knowledge, demonstrating that aligning the high-similar unpaired data is necessary.

\section{Conclusion}
\label{sec:conclusion}
In this work, we propose a novel skeleton-based training paradigm with vision-language knowledge prompts as the supervision rather than extensive human-designed coarse-grained annotations. Leveraging large multimodal models, we generate the vision-language knowledge prompts to establish the vision-language action concept space, enriching the skeleton action space with fine-grained details. In addition, we design the intra-modal self-similarity and inter-modal cross-consistency softened targets to effectively control and guide the degree to which samples should be selectively pulled closer in the instance discrimination and knowledge distillation process. These noisy pairs come from the abovementioned action space consisting of vision-language prompts and their corresponding skeleton. During the inference phase, only the skeleton serves as the input for action recognition. Experimental results show that our method achieves superior performance and promises better skeleton action representations.




\bibliography{IEEEabrv, reference}
\bibliographystyle{IEEEtran}

\begin{IEEEbiography}
[{\includegraphics[width=1in,height=1.25in,clip,keepaspectratio]{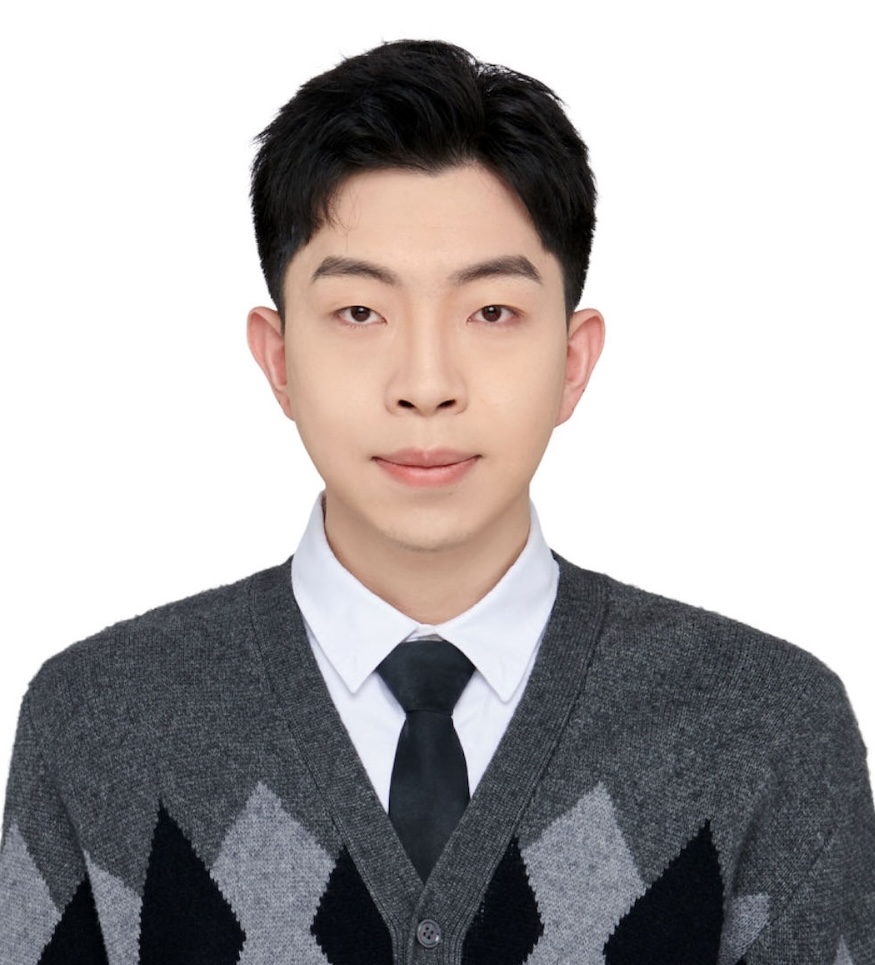}}]
{Yang Chen} received the M.Eng. degree in information and communication engineering from the School of Information and Communication Engineering at the University of Electronic and Technology of China (UESTC). Prior to that, he obtained a B.Sc. degree from the School of Physics and a B.B.A. degree from the School of Management and Economics, UESTC. He is currently pursuing a Ph.D. degree with the Department of Computing at The Hong Kong Polytechnic University (PolyU). His research interests include action recognition, zero-shot learning, and AI for Health.
\end{IEEEbiography}


\begin{IEEEbiography}
[{\includegraphics[width=1in,height=1.25in,clip,keepaspectratio]{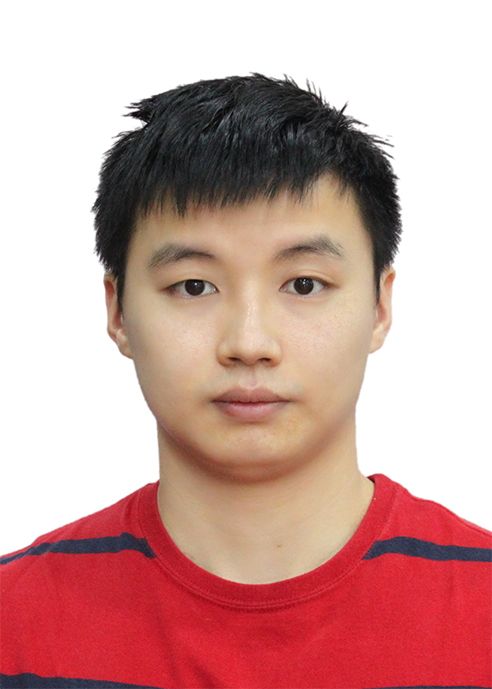}}]
{Tian He} received the B.E. degree from Southeast University, Jiangsu, China, in 2013 and the M.S. degree from the University of Pittsburgh, Pittsburgh, USA, in 2019. He is currently pursuing a Ph.D. degree in automation control engineering with the School of Automation at the University of Electronic and Technology of China (UESTC). His research interests include computer vision, action recognition, and action evaluation.
\end{IEEEbiography}


\begin{IEEEbiography}
[{\includegraphics[width=1in,height=1.25in,clip,keepaspectratio]{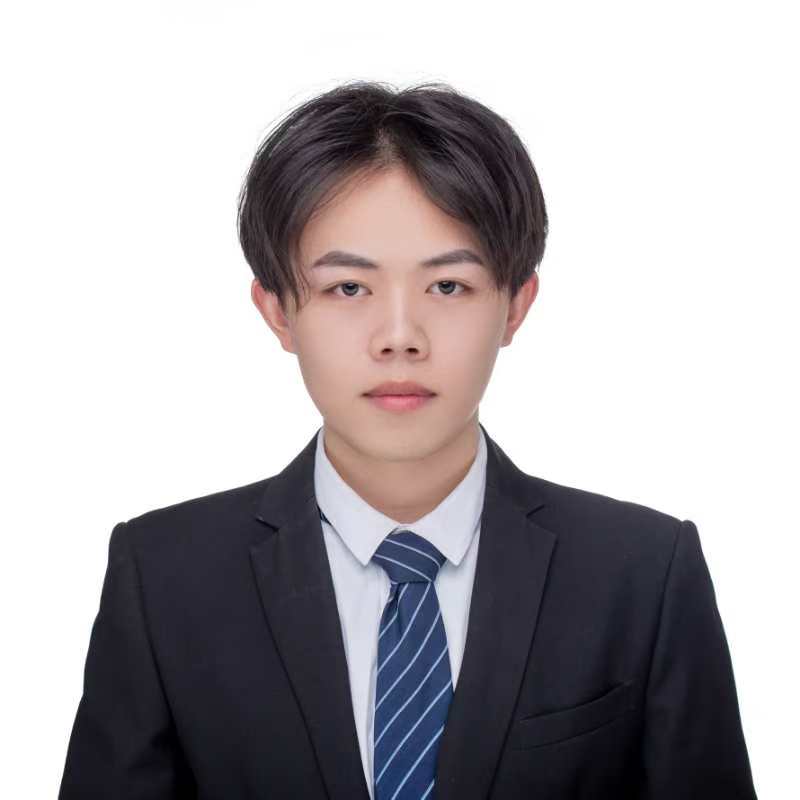}}]
{Junfeng Fu} received the B.E. degree in communication engineering from Wuhan University of Technology (WUT), Wuhan, China, in 2022. He is currently pursuing an M.Eng. degree in communication Engineering with the School of Information and Communication Engineering at the University of Electronic and Technology of China (UESTC). His research interests include automatic speech recognition and computer vision.
\end{IEEEbiography}


\begin{IEEEbiography}
[{\includegraphics[width=1in,height=1.25in,clip,keepaspectratio]{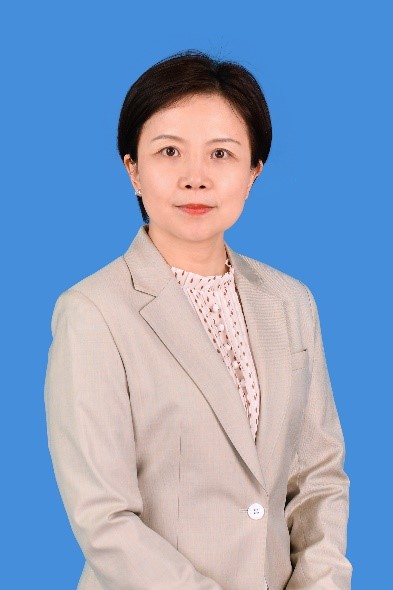}}]
{Ling Wang} (Member, IEEE) received the PhD. Degree in communication and system from University of Electronic Science and Technology of China in 2007. She works in University of Electronic Science and Technology of China from 2008, and currently is an associate professor with the school of communication and information engineering. She worked as a reach fellow at National University of Singapore from 2010 to 2011,and was a visiting scholar in University of Illinois at Urbana-Champaign from 2014 to 2015. Her main research interesting is pattern recognition and human-computer interaction.
\end{IEEEbiography}


\begin{IEEEbiography}
[{\includegraphics[width=1in,height=1.25in,clip,keepaspectratio]{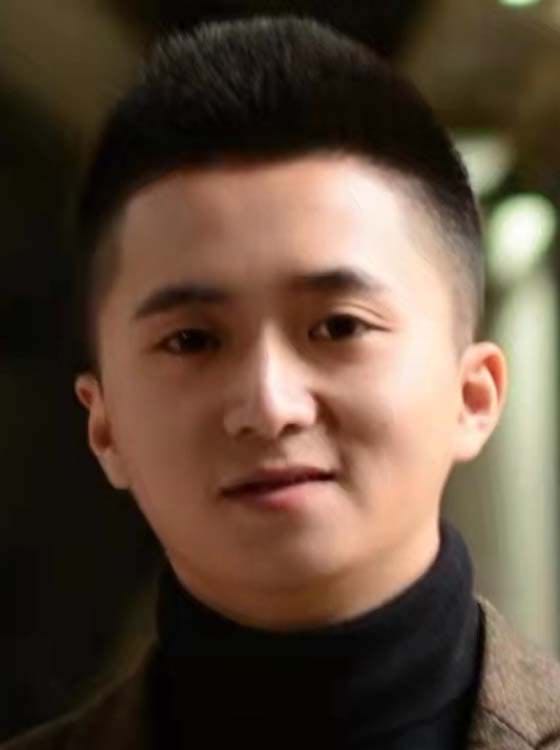}}]
{Jingcai~Guo} (Member, IEEE) is currently a Research Assistant Professor with Department of Computing, The Hong Kong Polytechnic University, Hong Kong SAR, where he received his Ph.D. in Dec. 2020. Prior to that, he received his M.E. from Waseda University (2015), Japan, and his B.E. from Sichuan University (2013), China, all in Computer Science. He is interested in Low-shot AI, which targets learning/modeling with limited resources in terms of data, computing capability, and their derivative applications. Topics include zero/few-shot learning, representation learning, federated learning, and model compression. He is currently serving as Associate Editor for IEEE Open Journal of the Computer Society and Guest Editor for IEEE Transactions on Computational Social Systems. He has served as Area Chair, Senior PC, and Session Chair for prestigious conferences like ICML, ACM-MM, AAAI, IJCAI, ICME, and VTC.
\end{IEEEbiography}


\begin{IEEEbiography}
[{\includegraphics[width=1in,height=1.25in,clip,keepaspectratio]{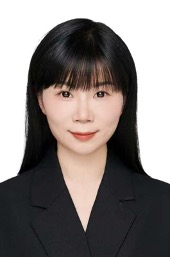}}]
{Ting Hu} is a Chief Physician and Director of the Rehabilitation Department at Chengdu First Orthopedic Hospital. She has extensive clinical experience in treating cervical spondylosis, lumbar disc herniation, shoulder pain, knee osteoarthritis, osteoporosis, various chronic pain conditions, as well as perioperative rehabilitation for trauma, joint and spinal fractures, rehabilitation for post-fracture functional disorders, assessment of adolescent non-idiopathic scoliosis, and integrated Chinese-Western medicine clinical diagnosis and treatment. She specializes in providing precise treatment plans that address both symptoms and root causes using ultrasound-guided techniques combined with modern rehabilitation therapies. 
\end{IEEEbiography}

\begin{IEEEbiography}
[{\includegraphics[width=1in,height=1.25in,clip,keepaspectratio]{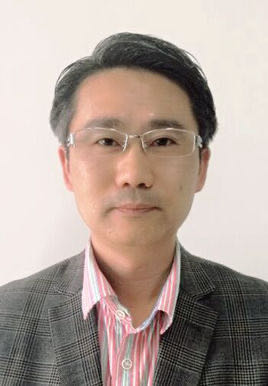}}]
{Hong Cheng} (Senior Member, IEEE) is a full Professor in School of Automation, a vice director of Center for Robotics, UESTC. He is the founding director of Machine Intelligence Institute, UESTC. He received Ph.D degree in Pattern Recognition and Intelligent Systems from Xi'an Jiaotong University in 2003. He was an associate Professor of Xi'an Jiaotong University since 2005. He was a postdoctoral at Computer Science School, Carnegie Mellon University, USA from 2006 to 2009. His current research interests include computer vision and machine learning, robotics. Dr.Cheng has been a senior member of IEEE, ACM, and Associate Editor of IEEE Computational Intelligence Magazine. Dr.Cheng serves as Finance Chair of ICME 2014, Local arrangement chair of VLPR 2012, Registration Chair of the 2005 IEEE ICVES.
\end{IEEEbiography}


\vfill

\end{document}